\documentclass[runningheads]{llncs}
\usepackage[utf8]{inputenc}
\usepackage{graphicx}
\usepackage{xcolor}
\usepackage{amssymb}
\usepackage{hhline}
\usepackage{multirow}

\definecolor{Cat}{HTML}{9F9BF9}

\color[rgb]{0,0,0} 

\definecolor{cgxi}{HTML}{C0DDEC}

\definecolor{cgxii}{HTML}{F4BEEE}

\newcommand{\variantof}[2]{{#1}_{#2}}

\newcommand{\textawareness}{structure-awareness}
\newcommand{\texthierinfo}{hierarchical composition information} %treelike components information ?

\title{SAN: Structure-Aware Network for Complex and Long-tailed Chinese Text Recognition}
\usepackage{caption}   % 图片脚注 
\usepackage{subfigure} % 子图包
\usepackage{float} 
\usepackage{graphicx}

\usepackage{algorithm}  
\usepackage{algorithmicx}  
\usepackage{algpseudocode}  
\usepackage{amsmath}  
\usepackage{pifont}
\usepackage{bbding}

\usepackage{hyperref}
\hypersetup{hidelinks,
	colorlinks=true,
	allcolors=black,
	pdfstartview=Fit,
	breaklinks=true}

\begin{document}
\author{Junyi Zhang\inst{1} \and Chang Liu \inst{1} \and Chun Yang\thanks{Corresponding Author}\inst{1}}
\institute{	School of Computer and Communication Engineering, University of Science and Technology Beijing \\
\email{zjy1926687175@163.com, lasercat@gmx.us, chunyang@ustb.edu.cn}
}
 
%\author{Junyi Zhang  ~~  Chang Liu ~~  Chun Yang*  University of Science and Technology Beijing}
%\institute{}
\maketitle

\begin{abstract}
%In text recognition, complex glyphs and long-tailed data have always been factors affecting model performance. Specifically for Chinese text recognition, characters with complex glyphs remain to be a main challenge for two reasons. First, due to the lack of shape-awareness, models tend to fail to model minor components of a complex character. Second, such characters appear less frequently in the training-set, making it harder for the model to capture its shape information. Hence in this work, we propose a structure aware network utilizing the hierarchical composition information to improve the recognition performance of complex characters. Implementation-wise, we first propose an auxiliary radical branch and integrate it into the base recognition network as a regularization term, which distills hierarchical composition information into the feature extractor. A Tree-Similarity-based weighting mechanism is then proposed to further utilize the depth information in the hierarchical representation. Experimental results demonstrate the proposed approach significantly improves the performances of complex characters, long-tailed characters, and the overall performance as well.

%In text recognition, complex glyphs and long-tailed data have always been factors affecting model performance.
In text recognition, complex glyphs and tail classes have always been factors affecting model performance.
%Specifically for Chinese text recognition, characters with complex glyphs remain to be a main challenge for two reasons. First, due to the lack of shape-awareness, models tend to fail to model minor components of a complex character. Second, such characters appear less frequently in the training-set, making it harder for the model to capture its shape information.
Specifically for Chinese text recognition, the lack of shape-awareness can lead to confusion among close complex characters. Since such characters are often tail classes that appear less frequently in the training-set, making it harder for the model to capture its shape information.
Hence in this work, we propose a structure-aware network utilizing the \texthierinfo{} to improve the recognition performance of complex characters. 
% , which boils down to the \textcompinfo{} and the \textstruinfo{},
% we propose to improve the performance of complex characters via improving the \textawareness{} of character features.
Implementation-wise, we first propose an auxiliary radical branch and integrate it into the base recognition network as a regularization term, which distills \texthierinfo{} into the feature extractor.
A Tree-Similarity-based weighting mechanism is then proposed to further utilize the depth information in the hierarchical representation.
% Since the \textcompinfo{} is shared by the tail and head classes alike, 
%Experimental results demonstrate the proposed approach significantly improves the performances of complex characters, long-tailed characters, and the overall performance as well.
Experiments demonstrate that the proposed approach can significantly improve the performances of complex characters and tail characters, yielding a better overall performance.
Code is available at \href{https://github.com/Levi-ZJY/SAN}{https://github.com/Levi-ZJY/SAN}
%and the overall Chinese text recognition task as well. 


%In text recognition, complex glyphs and long-tailed data have always been factors affecting model performance.
%For Chinese text recognition, challenges like the large character set, diversity of fonts, complex glyphs, and long-tailed data distribution yield particularly prominent negative effects on the recognition performance. Character \textawareness{} can alleviate the confusion among alike complex characters.
%Hence in  this work, we propose to improve the performance of complex characters via improving the \textawareness{} of character features.
%Specifically, we propose to exploit the \texthierinfo{} of each character, which boils down to the \textcompinfo{} and the \textstruinfo{}.  
%The proposed approach also improves the performance of tail classes that less frequently appear in training data. 
%Implementation-wise, we propose an auxiliary radical branch and integrate it into the base recognition network as a regularization term, to distill hierarchical components information into the feature extractor. Experimental results demonstrate the module significantly improves the performances of complex characters long-tailed characters, and the overall Chinese text recognition task as well. 

%\footnote{* The corresponding author}

%\input{abstract_v1}
\end{abstract}

\section{Introduction}
%Part1:
%（why doing this）
Chinese text recognition plays an important role in the field of text recognition due to its huge audience and market. 
%However, the performances of the Chinese text recognition \cgone{methods} have often been limited by the large scale \cgone{of} character vocabulary, the large number of complex characters, and the insufficiency of training samples~\cite{ZeroshotHC}.
%\cgone{However, the performances have often been limited due to several traits of the Chinese language, including the large character set, the large number of complex characters, and the sample scarcity on tailed characters. ~\cite{ZeroshotHC}}
%Part2:
%(shortcomings of the existing methods)
Most current text recognition methods, including Chinese text recognition methods, are character based, where characters are the basic elements of the prediction. Specifically, most methods fit into the framework formulated by Beak et al.~\cite{www}, which includes an optional rectifier (Trans.), a feature extractor~(Feat.), a sequence modeler~(Seq.), and a character classifier~(Pred.). Many typical Chinese text recognition methods~\cite{MultiColumnDN,HandwrittenCR,ICDAR2C}, also fall into this category, where the feature extractor generally takes the form of a Convolutional Neural Network, and the classifier part is mostly implemented as a linear classifier decoding input features into predicted character probabilities. However, the naive classification strategy has limited performance on Chinese samples, due to the large character set, severely unbalanced character frequency, and the complexity of Chinese glyphs.

To address the frequency skew, compositional learning strategies are widely used in low-shot Chinese character recognition tasks~\cite{RANRA,taktak,wubizhengma}. For compositional information exploited, 
%the majority of implementations~\cite{RANRA,RadicalAN,ZeroshotHC,DenseRANFO,JointSA} utilize the radical sequence of each character, where the components and the composition structures can be hierarchically modeled as a tree.
the majority of implementations~\cite{RANRA,RadicalAN,ZeroshotHC,DenseRANFO,JointSA} utilize the radical 
representation, where the components and the structural information of each character are hierarchically modeled as a tree. Specifically,
the basic components serve as leaf nodes and the structural information (the spatial placement of each component) serves as non-leaf nodes. 
Some methods are also seen to utilize stroke~\cite{taktak} or Wubi~\cite{wubizhengma} representations. Besides the Chinese language, characters in many other languages can be similarly decomposed into basic components~\cite{taktak,9412607,Rai2021PhoSCNetAA,DBLP:conf/icfhr/ChandaBHHSS18}. These methods somewhat improve text recognition performance under low-shot scenarios. However, compositional-based methods are rarely seen in regular recognition methods due to their complexity and less satisfactory performance. 
This limitation is solved by PRAB~\cite{fudanvi}, which proposes to use radical information as a plug-in regularization term. The method decodes the character feature at each timestamp to its corresponding radical sequence and witnesses significant overall performance improvement on several SOTA baselines. However, the method still has two major limitations. First, PRAB~\cite{fudanvi} only applies to text recognition methods with explicitly aligned character features~\cite{ASTER,SAR} and does not apply to implicitly aligned CTC-based methods like CRNN~\cite{CRNN}. Furthermore, PRAB and most of the aforementioned radical method treats large parts and small parts alike, ignoring the depth information of the hierarchical representation.

%In this work,
To alleviate the limitations, we propose a Structural-Aware Network~(SAN) which distills \texthierinfo{} into the feature extractor with the proposed alignment-agnostic and depth-aware Auxilary Radical Branch (ARB). 
%ARB serves as a regularization term that 
%allows the radical level prediction model to supplement and 
%refines the visual feature extraction of the character level prediction model in the training process,
The ARB serves as a regularization term which directly refines feature maps extracted by the Feat. stage to preserve the \texthierinfo{} without explicitly aligning the feature map to each character. 
The module allows the model to focus more on local features and learn more structured visual features, which
significantly improves complex character recognition accuracy. As basic components are shared among head and tail classes alike, it also improves the tail-classes performance by explicitly exploiting their connections with head classes.
%Furthermore, we proposed a novel Tree Similarity~(TreeSim) method that serves as a more fine-grained metric measuring the visual similarity between two characters. 
Furthermore, we proposed a novel Tree Similarity~(TreeSim) semimetric serves as a more fine-grained measure of the visual similarity between two characters. 
%Based on the new metric, we propose to further
The proposed TreeSim semimetric further allows us to exploit the depth information of the hierarchical representation, which is implemented by weighting the tree nodes accordingly in ARB.
%choose to validate the aforementioned assumption on the Chinese language, and the %hierarchical radical representation~\cite{hde} is adopted to model the structural and component information of each character.  
%Based on the radical decomposition method, we propose a strategy to enhance the \textawareness{} of visual features by using the hierarchical representation as a regularization term. 
%The proposed approach significantly improves complex character recognition accuracy. As basic components are shared among head and tail classes alike, it also improves the tail-classes performance by explicitly exploiting their connections with head classes.
The suggested method substantially enhances the accuracy of complex character recognition. As fundamental elements are shared between head and tail classes, it also boosts the performance of tail classes by leveraging their relationships with head classes.

Experiments demonstrate the effectiveness of ARB in optimizing the recognition performance of complex characters and long-tailed characters, and it also improves the overall recognition accuracy of Chinese text.
The contributions of this work are as follows:
\begin{itemize}
    \item We propose a SAN for complex character recognition by utilizing the hierarchical components information of the character.

    \item ARB based on the tree modeling of the label is introduced, which enhances the structure awareness of visual features. 
    ARB shows promising improvement in complex character and long-tailed character recognition and it also improves the overall recognition accuracy of Chinese text.

    \item We propose a novel TreeSim method to measure the similarity of two characters, 
    %which is hierarchical, bidirectional, and deep-independent.
    and propose a TreeSim-based weighting mechanism for ARB to further utilize the depth information in the hierarchical representation.

\end{itemize}

\section{Related Work}

\subsection{Character-based Approaches} 
In Chinese text recognition, early works are often character-based. Some works are based on CNN model~\cite{MultiColumnDN,HandwrittenCR,ICDAR2C} to design improved or integrated methods. For example, MCDNN~\cite{MultiColumnDN} integrates multiple models including CNN, which shows advantageous performance in handwritten characters recognition. ART-CNN~\cite{HandwrittenCR} alternatively trains a relaxation CNN to regularize the neural network during the training procedure and achieves state-of-the-art accuracy. Later, DNN-HMM~\cite{DeepNN} sequentially models the text line and adopts DNN to model the posterior probability of all HMM states, which significantly outperforms the best over-segmentation-based approach~\cite{HandwrittenCT}. 
The SOTA text recognition model, ABINet~\cite{ABINet}, recommends blocking the gradient flow between the vision and language models, and introduces an iterative correction approach for the language model. These strategies promote explicit language modeling and effectively mitigate the influence of noisy input.
However, these methods did not put forward effective countermeasures against the difficult problems of Chinese text recognition like many complex characters, insufficient training samples, and so on, so the performance improvement of these models is greatly limited.

\subsection{Chinese enhanced Character-based Approaches} 

Several methods attempt to design targeted optimization strategies according to the characteristics of Chinese text~\cite{FromTT,MaximumER,WriterAwareCF,TemplateInstanceLF,JointAA,BuildingFA}. Wu et al~\cite{FromTT}. use MLCA recognition framework and new writing-style-aware image synthesis method to overcome large character sets and great insufficient training samples problems. In~\cite{MaximumER}, the authors apply Maximum Entropy Regularization to regularize the training process to optimize the large amount of fine-grained Chinese characters and the great imbalance over class problems. Wang et al.~\cite{WriterAwareCF} utilize the similarity of Chinese characters to reduce the total number of HMM states and model the tied states more accurately. These methods pay attention to the particularity of Chinese characters and give targeted optimization. However, they are still character-based, which makes it difficult to further explore the deep features of Chinese characters.

\subsection{Radical-based Approaches}

In recent years, radical-based approaches have shown outstanding advantages in Chinese recognition~\cite{RANRA,RadicalAN,ZeroshotHC,DenseRANFO,JointSA}. RAN~\cite{RANRA} employs an attention mechanism to extract radicals from Chinese characters and to detect spatial structures among the radicals, which reduce the vocabulary size and can recognize unseen characters. FewShotRAN~\cite{RadicalAN} maps each radical to a latent space and uses CAD to analyze the radical representation. HDE~\cite{ZeroshotHC} designs an embedding vector for each character and learns both radicals and structures of characters via a semantic vector, which achieves superior performance in both traditional HCCR and zero-shot HCCR tasks. Inspired by these works, our method emphasizes the role of structural information of radical in visual perception and proposes to utilize the common local and structural traits between characters to optimize the recognition performance in complex and long-tailed characters.

\section{Our Method}
\begin{figure}[t]
  \includegraphics[width=\linewidth]{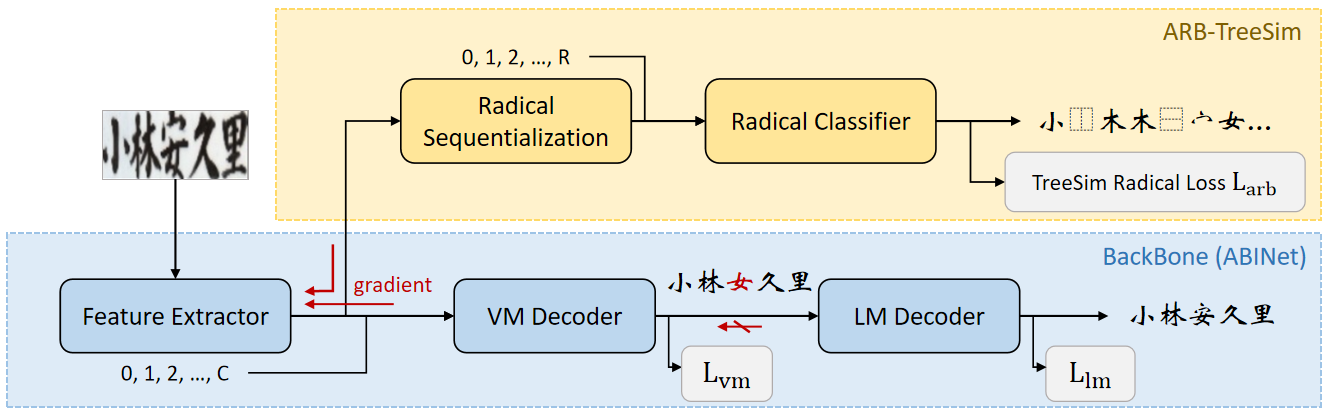}
  \caption{The Structure-Aware Network (SAN). The orange square frame is the ARB-TreeSim and the blue square frame is the base network. The gradient flow of the ARB-TreeSim and the VM decoder influence the feature extractor together. }
  \label{fig:framewoek}
\end{figure}
In this work, we propose the Structure-Aware Network (SAN, Fig.~\ref{fig:framewoek}) which composes a base method~(in blue) and a proposed Auxiliary Radical Branch~(ARB). 
Implementation-wise, we adopt the SOTA ABINet~\cite{ABINet} as an example base method. ABINet features separate vision and language models, with no gradient flow occurring between them. The optimization of the vision model and the language model, denoted as  $L_{vm}$ and $L_{lm}$ respectively, is carried out independently from each other. Specifically, the ARB is applied to the sample feature map extracted by the Vision Model, yielding a novel two branches learning network. During training, the feature map is processed by both branches, one is passed into the VM decoder, decoded by characters, and the other is passed into ARB, decoded by radical. During evaluation, the model reduces into the corresponding base model, yielding no extra costs. 

The following part of this section first introduces the hierarchical representation of characters and the Tree Similarity~(TreeSim) metric that measure the similarity of two radical trees.  Then we introduce the proposed Auxiliary Radical Branch, which is a regularization term to improve the model shape-awareness by distilling the hierarchical representation. We then introduce the TreeSim enhanced Weighting which adds depth information to the ARB branch.  
%At last, we integrate the proposed model into a base recognition network and demonstrate its efficacy. 
\subsection{Label Modeling } 

As characters can be decomposed into basic components, 
we model it as a tree composed of various components hierarchically and model the label as a forest composed of character trees, which gives the label a structural representation.
%This representation can improve the \textawareness{} of the vision model and allows it to pay more attention to the details and local features of characters instead of the overall information.

 As shown in Fig.2~(a), in Chinese, one popular modeling method is to decompose characters into radicals and structures~\cite{fudanvi}. 
 Radicals are the basic components of characters, and structures describe the spatial composition of the radicals in each character.
 One structure is often associated with two or more radicals, and the structure is always the root node of a subtree.
 %we can regard it and the associated radicals as a subtree.
 %The subtree can be regarded as a big radical, which is connected to another structure to form a larger subtree. 
 In this way, all Chinese characters can be modeled as a tree, in which leaf nodes are radicals, and non-leaf nodes are structures. For simplicity, we call such trees ``radical trees'' in this paper.

%\begin{figure}[t]	
%    \begin{minipage}[t]{0.5\linewidth}
%       \centering          %子图居中
%        \includegraphics[height=3cm,width=6.2cm]{figures/RadicalTree.png}   
%        \caption{Radical Tree. The red nodes are called structure and the blue nodes are called radical.}%The right side is the result of converting the left radical structure sequence into a tree.}
%    \end{minipage}
%    \begin{minipage}[t]{0.5\linewidth}
%        \centering          %子图居中
%    \end{minipage}
%   \begin{minipage}[t]{0.5\linewidth}
%        \centering          %子图居中
%    \end{minipage}
%    \begin{minipage}[t]{0.5\linewidth}
%        \centering          %子图居中
%    \end{minipage}
%    \begin{minipage}[t]{0.5\linewidth}
%        \centering      %子图居中
%        \includegraphics[height=3.5cm,width=4.5cm]{figures/TreeSimWeight.png}
%        \caption{TreeSim Weight. The yellow circle indicates that the total weight of the subtree is 1/3.}
%    \end{minipage}
%\end{figure}

\begin{figure}[t]	
\centering
    \subfigure[Radical Tree]
    {
        \begin{minipage}[t]{0.45\linewidth}
            \centering          %子图居中
            \includegraphics[height=3cm,width=6.2cm]{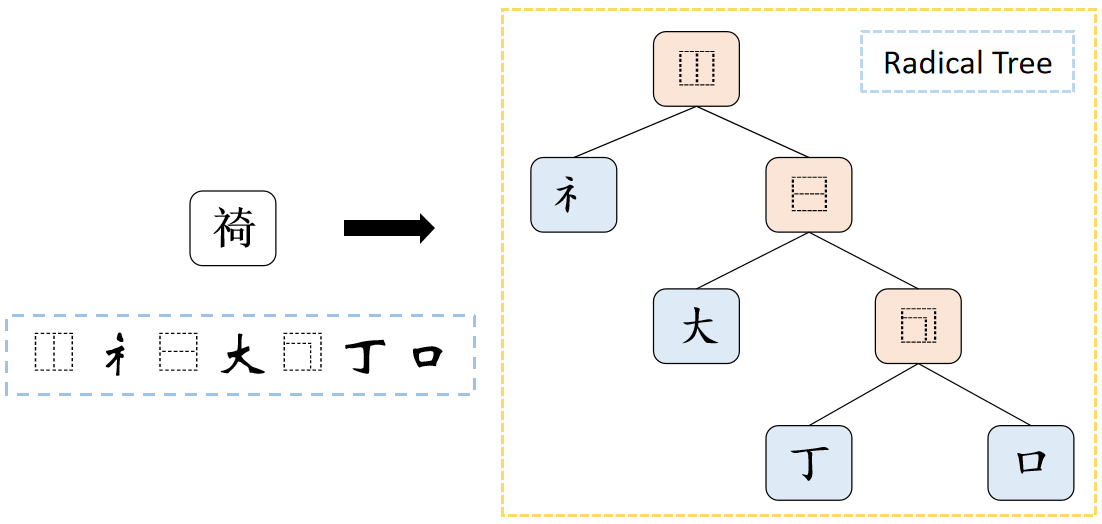}   
            %\caption{Radical Tree}
        \end{minipage}
    }
    \subfigure[TreeSim Weight]
    {
        \begin{minipage}[t]{0.45\linewidth}
            \centering          %子图居中
            \includegraphics[height=3.5cm,width=4.5cm]{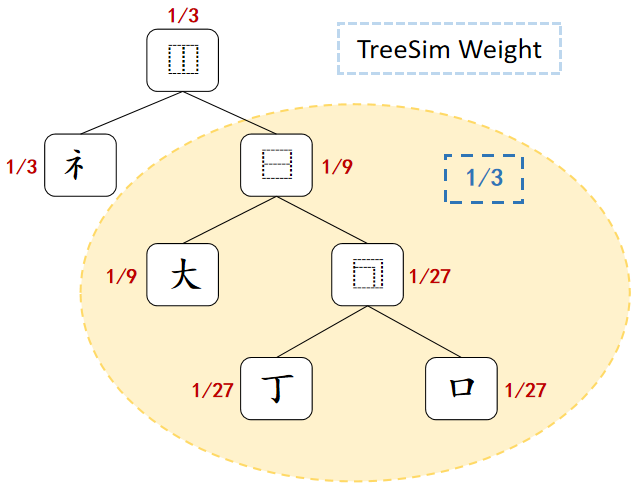}   
            %\caption{Radical Tree}
        \end{minipage}
    }
\caption{In figure (a), the red nodes are called structure and the blue nodes are called radical. In figure (b), the yellow circle indicates that the total weight of the subtree is 1/3.}
\end{figure}

In this work, we propose a novel Tree Similarity~(TreeSim) metric to 
measure the visual similarity between two characters represented with radical trees.
In a radial tree, the upper components represent large and overall information, while the lower components represent small and local information. When comparing the similarity of two characters, humans tend to pay more attention to the upper components and less attention to the lower ones.

% \begin{algorithm}[!h]\small
%     \caption{TreeSim Weight Algorithm}  
%     \begin{algorithmic}[1]
%         %\Require 
%         %\Ensure   
%         %{\textbf{Input:}}
%         \Require  
%           $Node$: Pointer of the root node;  
%           $W$: Weight of the tree, which is equal to 1;
%         \Function {GetTreeSimWeight}{$Node, W$}  
%             %\State $n \gets $number of branches of this node  
%             \State $n \gets $ degree of this node  
%             \If {$n == 0$}  
%                 \State weight of this node$\gets W$
%                 \State \Return{}
%             \Else
%                 \If {$n == 2$}  
%                     \State weight of this node$\gets W/3$
%                     \State \Call{GetTreeSimWeight}{$Node.leftchild, W/3$}  
%                     \State \Call{GetTreeSimWeight}{$Node.rightchild, W/3$}
%                 \EndIf
%                 \If {$n == 3$}  
%                     \State weight of this node$\gets W/4$
%                     \State \Call{GetTreeSimWeight}{$Node.leftchild, W/4$}
%                     \State \Call{GetTreeSimWeight}{$Node.midchild, W/4$}
%                     \State \Call{GetTreeSimWeight}{$Node.rightchild, W/4$}
%                 \EndIf
%             \EndIf
%             \State \Return{}  
%         \EndFunction  
%     \end{algorithmic}  
% \end{algorithm}  

Correspondingly, as shown in Fig.2~(b), we first propose a weighting method for TreeSim. The method includes the following three characteristics:
%Correspondingly, Tree-edit Distance includes the following characteristics:
First, upper components in the radical tree have greater penalty weight. Second, every subtree is regarded as an independent individual, the root node and its subtrees have equal weight. Third, the total weight of the tree is 1.
The calculation procedure is depicted in Algorithm 1, where $Node$ is the pointer of the root node and $w_{sub}$ is the weight of the subtree. The weight of each node can be calculated by the recursion method.

\begin{algorithm}[t]\small
    \caption{TreeSim Weight Algorithm}  
    \begin{algorithmic}[1]
        %\Require 
        %\Ensure   
        %{\textbf{Input:}}
        \Require  
          $Node$: Pointer of the root node;  
          
          ~~~~$w_{sub}$: Weight of the subtree, which equals to 1 in the initial call;
        \Function {GetTreeSimWeight}{$Node, w_{sub}$}  
            %\State $n \gets $number of branches of this node  
            \State $n \gets $number of children of this node  
            \If {$n == 0$}  
                \State weight of this node$\gets w_{sub}$
                \State \Return{}
            \Else
                \State weight of this node$\gets w_{sub}/(n+1)$
                \For{$c \gets Node.Children$}                    
                    \State \Call{GetTreeSimWeight}{$c, w_{sub}/(n+1)$}  
                \EndFor
            \EndIf
            \State \Return{}  
        \EndFunction  
    \end{algorithmic}  
\end{algorithm}

%2)The subtree and its root node are considered as a whole, the weight distributed to the root node is distributed to the whole. 3)The total weight of the whole tree is 1, the root node and its directly connected nodes are equally distributed with the root node weight.

Based on this weighting method, the TreeSim calculation sample is shown in Fig.3. 
The calculation process is as follows: 
First, select any one of the two radical trees, and traverse every node in preorder.
Second, judge whether it is matching for each node. The matching rules include: 1)Every ancestor of this node is matching. 2)This location in the other tree also has a node and the two nodes are the same. 
Last, TreeSim is equal to the sum of the weights of all matching nodes.

The proposed TreeSim is hierarchical, bidirectional, and deep-independent. 
Hierarchical means TreeSim pays different attention to different levels of the tree.
Bidirectional means no matter select which of the two trees, the calculation result is the same. 
Depth-independent means the weight of nodes at a certain level is independent of the depth of the tree.

Based on this label modeling method, we proposed two loss design strategies.
Considering that a rigorous calculation for the loss of two forests may be complex and nonlinear, we introduce two linear approximate calculation methods in 3.2 to supervise the loss calculation.

\begin{figure}[t]	
    \begin{minipage}[b]{1\linewidth}
        \centering          %子图居中
        \includegraphics[height=6.5cm,width=11cm]{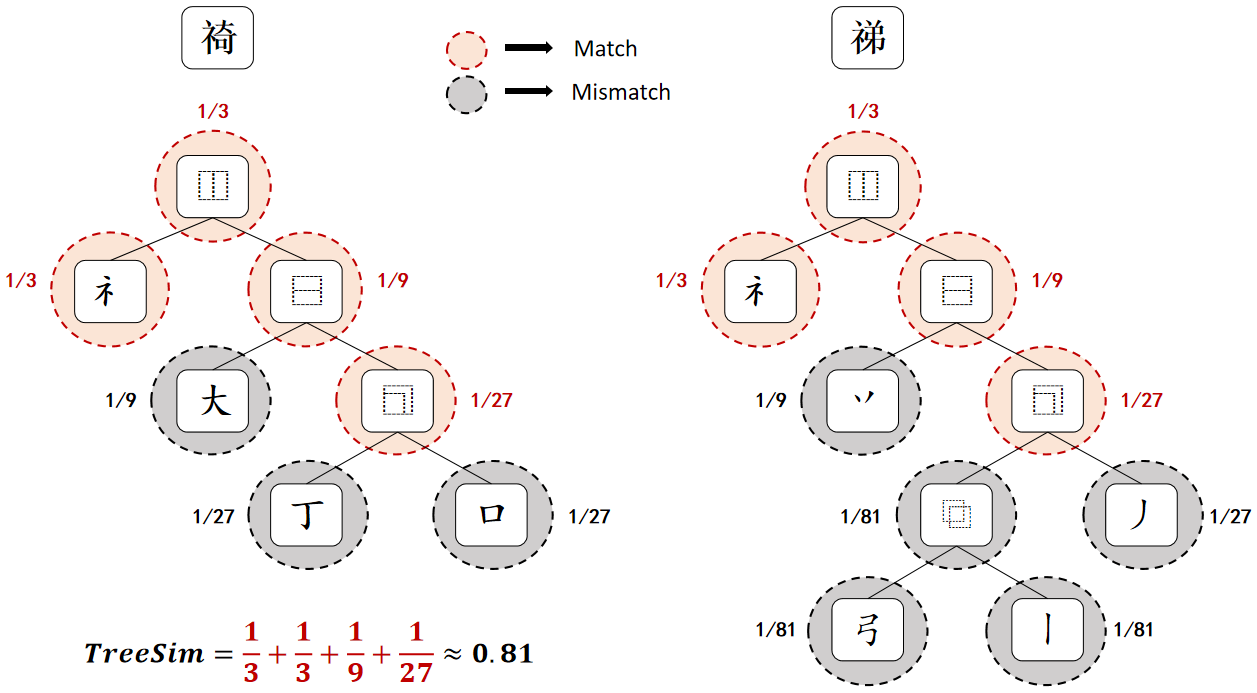}
        \caption{TreeSim Calculation. Nodes with a red circle are the match modes and nodes with a black circle are the mismatch nodes. The calculation results of the two radical trees are the same.}
    \end{minipage}
    %\vspace{-7mm}
\end{figure}

\subsection{Auxiliary Radical Branch} 

Inspired by the radical-based methods~\cite{RadicalAN,fudanvi}, we first propose an Auxiliary Radical Branch~(ARB) module to enhance the \textawareness{} of visual features by using the hierarchical representation as a regularization term. Unlike the PRAB~\cite{fudanvi} method, the proposed ARB module directly decodes the feature map extracted from the Feat. stage, thus do not need individual character features and can be theoretically applied to most methods with a Feat. stage~\cite{www}. 
As shown in Fig.1, ARB includes the following characteristics: 1)The label representation in ARB is hierarchical, which contains component and structural information. 2)ARB is regarded as an independent visual perception optimization branch, which takes feature as input, decodes according to structured representation, and finally refines the visual feature extraction.

The role of ARB is reflected in two aspects: One is that ARB can use hierarchical label representation to supply and refine the feature extraction, improving the \textawareness{} of the model.
%push the model to pay more attention to the local features and learn more structured visual representation. 
The other is that by taking advantage of the common component combination traits between Chinese characters, ARB could exploit the component information connections between tail classes and head classes, which can optimize long-tailed character recognition.
%ARB could transfer the learning results from different characters, and optimize the long-tailed character recognition.

In ARB, we propose two linear loss design strategies, which reflect the hierarchical label representation while simplifying the design and calculation.

\subsubsection{Sequence Modeling}

%The Chinese label is hierarchically modeled as a forest (discussed in 3.1), 
Considering that the proposed label modeling method is top-down and root-first, in loss design, we intend to perpetuate this design mode to match them.
We propose using the sequence by traversing the radical tree in preorder to model each tree in the label, so we use the linear radical structure sequence~\cite{fudanvi}.
%We propose to model the label by concatenating the pre-order traverse of the component tree of each character. Implementation-wise,  we adopt the character representation in ~\cite{fudanvi}. 
This sequence follows the root-first design strategy and retains the hierarchical structure information of the radical tree to a certain extent. 
By connecting each radical structure sequence in the label, we get the sequence modeling label.

%for each tree in the forest, we use the linear radical structure sequence~\cite{fudanvi} by traversing the tree in preorder.
%This sequence retains the hierarchical structure information of Chinese character components to a certain extent. By connecting each radical structure sequence in label, we get the sequence modeling label.

The ARB prediction is supervised with the ARB loss $L_{arb}$, which is a weighted cross-entropy loss of each element.
\begin{equation}
L_{arb}=\sum_{i}^{\variantof{l}{rad}}w_{rad_{(i)}}*logP(r^{*}_{(i)}),
\end{equation}
where $r^{*}_{i}$ is the ground truth of the $i$-th radical, $l_{rad}$ is the total radical length of \textbf{all} characters in the sample.

The radical structure sequence contains the component and structural information of the label, so this modeling method implements the introduction of hierarchical information and effectively improves the vision model performance.
\subsubsection{Naive Weighting}
We first use a naive way to set equal weights to radicals in each character, i.e, $\mathbf{w}_{rad}=\mathbf{w}_{naive}= \mathbf{1}.$
Albeit the Naive method demonstrates some extent of performance improvement, it treats all components as equal importance, which may lead to over-focusing on minor details.

\subsubsection{TreeSim Enhanced Weighting}

In sequence modeling, the structural information is implicit, we want to make it more explicit. 
%we want to better model the hierarchical structure of radicals. 
%So we propose to use Tree-edit Distance to calculate the difference between two radial trees. Based on it, 
So we propose TreeSim enhanced Weighting method to explicitly strengthen the structural information of label representation.

We add the TreeSim weight~(Fig.2(b)) to the naive weight as a regularization term and get the final radical weight.
\begin{equation}
    \mathbf{w}_{rad}= \mathbf{w}_{naive} + \lambda_{TreeSim}\mathbf{w}_{TreeSim}
    %w_{radical}= \frac{{}w_{soft} + \lambda_{ted}w_{ted}}{2}
    %w_{radical}= (1-\lambda{})w_{soft} +  \lambda{}\mathbf{w}_{ted},
\end{equation}
where $\lambda_{TreeSim}$ is set to 1.
This method enhanced the structural information in loss explicitly, which shows better performance on the vision model than the Naive method.

\subsection{Optimization} 
%We combine the module with the SOTA ABINet to demonstrate the efficacy of the proposed ARB module. ARB is added behind the feature extractor. The feature is divided into two branches, one is passed into the VM decoder, decoded by characters, and the other is passed into ARB, decoded by radical.
The training gradient flow by two branches is superimposed and updates the weight of the feature extractor together, which allows the radical level prediction model to supplement and refine the visual feature extraction of the character prediction model in the training process.
%allowing them to update the weight of the feature extractor together.
we define the loss as follows
\begin{equation}
    L_{overall}=L_{base}+L_{arb},
\end{equation}
where $L_{base}$ is the loss of base model, which is the loss from ABINet in this work, i.e.,
\begin{equation}
    L_{base}=L_{vm}+L_{lm}.
\end{equation}

Both the character prediction branch and radical structure prediction branch affect the feature extraction. For fairness, we give them the same weight of influence on visual feature extraction, to ensure that the performance of both parties is fully reflected.

%\begin{figure}[t]	
%    \begin{minipage}[b]{1\linewidth}
%       \centering       
%        \includegraphics[height=2.3cm,width=12cm]{figures/ExampleResult2.png}   
%        \caption{Successful recognition examples using ARB. The two pictures on the left are complex character examples. The two pictures on the right are long-tailed character examples. Text strings are ABINet prediction and ARB-TreeSim prediction}
%    \end{minipage}
%    \vspace{-7mm}
%\end{figure}

%\input{ourmethod_if_cat.tex}

%\input{experiments}
%\input{experiments_change.tex}
\section{Experiment}

\subsection{Implementation Details}
The datasets used in our experiments are Web Dataset and Scene Dataset~\cite{fudanvi}. The Web Dataset contains 20,000 Chinese and English web text images from 17 different categories on the Taobao website and the Scene Dataset contains 636,455 text images from several competitions, papers, and projects.
The number of radical classes is 960.
We set the value of R~(Fig.1) to be 33 for the web dataset and 39 for the scene dataset, covering a substantial 95$\%$ of the training samples from both datasets.

We implement our method with PyTorch and conduct experiments on three NVIDIA RTX 3060 GPUs. 
Each input Image is resized to $32\times128$ with data augmentation.
The batch size is set to 96 and ADAM optimizer is adopted with the initial learning rate of $1e^{-4}$, which is decayed to $1e^{-5}$ after 12 epochs.

\subsection{Ablative Study}
\begin{table}[b]
    \caption{Ablation study of ARB. The dataset used is Web dataset~\cite{fudanvi}.}

    \centering
    \begin{tabular}{|c|c|c||c|c|}
        \hline
         Model& Naive Radical loss & TreeSim weighing loss & Accuracy & 1-NED \\
         \hline
         VM-Base& & & 59.1 & 0.768 \\
         VM-Naive& \checkmark & & 60.7 & 0.779 \\
         \textbf{VM-TreeSim} & \checkmark &\checkmark& \textbf{61.3} & \textbf{0.786} \\
         \hline
         ABINet-Base & & & 65.6 & 0.803  \\
         ABINet-Naive &\checkmark & & 66.8 & 0.812  \\
         \textbf{ABINet-TreeSim (SAN)}  &\checkmark &\checkmark & \textbf{67.3} & \textbf{0.817} \\
         \hline
    \end{tabular}
    \label{tab:my_label}
\end{table}

We discuss the performance of the proposed approaches~(ARB and TreeSim weighting) with two base-model configurations: Vision Model and ABINet. Experiment results are recorded in Tab.1. 
%From the statistics we can \cgone{draw the following two conclusions.}

%First, the proposed ARB is useful, which enhances the accuracy in both VM and ABINet.
First, the proposed ARB is proved useful, by significantly improving the accuracy of both base-metods. Specifically, the
VM-Naive outperforms VM-Base by 1.6$\%$ accuracy and 0.011 1-NED, ABINet-Naive outperforms ABINet-Base by 1.2$\%$ accuracy and 0.009 1-NED. 
Second, TreeSim enhanced weighting also achieves expected improvements. VM-TreeSim boosts accuracy by 0.6$\%$ and  1-NED by 0.007 than VM-Naive, SAN boosts accuracy by 0.5$\%$ and 1-NED by 0.005 than ABINet-Naive.

The above observations suggest that the hierarchical components information is useful to the feature extractor, which can significantly improve the performance of the vision model. The proposed approaches also yields significant improvement against full ABINet indicating hierarchical components information still plays a significant part, even when the language models can, to some extent, alleviate the confusions caused by insufficiency shape awareness.

\subsection{Comparative Experiments}
%1) compare with sota(fudanvi).

\begin{table}[t]
    \caption{Performance on Chinese text recognition benchmarks~\cite{fudanvi}. $\dagger$ indicates results reported by ~\cite{fudanvi}, $*$ indicates results from our experiments. }

    \centering
    \begin{tabular}{|c||c|c|c|c|}
        \hline
         Method & \multicolumn{2}{c|}{Web} & \multicolumn{2}{c|}{Scene}  \\
         \hhline{~----}
         & Accuracy  &1-NED &Accuracy&1-NED\\
         \hline
         CRNN$\dagger$~\cite{CRNN} & 54.5&0.736 & 53.4&0.734   \\
         ASTER$\dagger$~\cite{ASTER} & 52.3&0.689 & 54.5&0.695  \\
         MORAN$\dagger$~\cite{moran} & 49.9&0.682 & 51.8&0.686  \\
         SAR$\dagger$~\cite{SAR} & 54.3&0.725 & 62.5&0.785 \\
         SRN$\dagger$~\cite{SRN} & 52.3&0.706 & 60.1&0.778   \\
         SEED$\dagger$~\cite{seed} & 46.3&0.637 & 49.6&0.661  \\
         TransOCR$\dagger$~\cite{Chen2021SceneTT} & 62.7&0.782 & 67.8&0.817 \\
         TransOCR-PRAB$\dagger$~\cite{fudanvi} & 63.8&0.794 & 71.0 &0.843 \\ 
         ABINet$*$~\cite{ABINet} & 65.6 & 0.803 & 71.8 & 0.853 \\
         \hline
         SAN (Ours) & \textbf{67.3} & \textbf{0.817} & \textbf{73.6} & \textbf{0.863} \\
         \hline
    \end{tabular}
    \label{tab:my_label}
\end{table}

Compared with Chinese text recognition benchmarks and recent SOTA works that are trained on web and scene datasets, SAN also shows impressive performance(Tab.2). We can see from the comparison, our SAN outperforms ABINet with 1.7$\%$, 1.8$\%$ on Web and Scene datasets respectively. Also, SAN achieves the best 1-NED on both datasets. Some successful recognition examples are shown in Fig.4, which shows the complex characters and long-tailed characters that predict failure using the base model (ABINet) while predicted successfully by the full model.

\begin{figure}[t]	
\centering
    \subfigure[Complex]
    {
        \begin{minipage}[t]{0.22\linewidth}
            \centering          %子图居中
            \includegraphics[height=2.2cm,width=3cm]{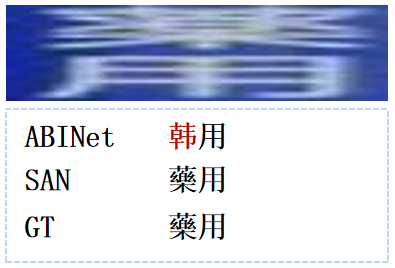}   
            %\caption{Radical Tree}
        \end{minipage}
    }
    \subfigure[Complex]
    {
        \begin{minipage}[t]{0.22\linewidth}
            \centering          %子图居中
            \includegraphics[height=2.2cm,width=3cm]{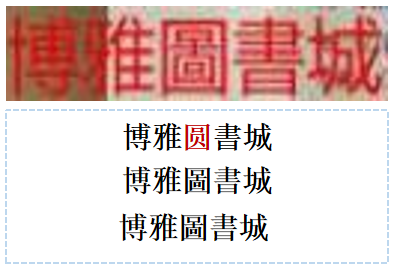}   
            %\caption{Radical Tree}
        \end{minipage}
    }
    \subfigure[Long-tailed]
    {
        \begin{minipage}[t]{0.22\linewidth}
            \centering          %子图居中
            \includegraphics[height=2.2cm,width=3cm]{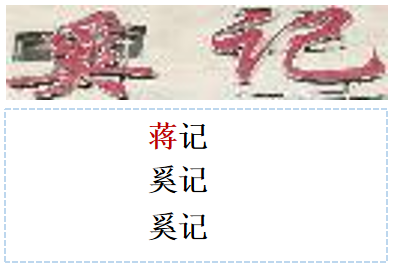}   
            %\caption{Radical Tree}
        \end{minipage}
    }
    \subfigure[Long-tailed]
    {
        \begin{minipage}[t]{0.22\linewidth}
            \centering          %子图居中
            \includegraphics[height=2.2cm,width=3cm]{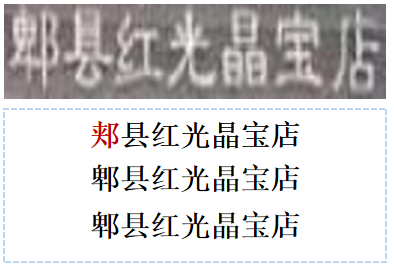}   
            %\caption{Radical Tree}
        \end{minipage}
    }
\caption{Successful recognition examples using ARB. (a) (b) are complex character examples. (c) (d) are long-tailed character examples.The text strings are ABINet prediction, SAN prediction and Ground Truth.}
\end{figure}

\subsection{Property analysis}
%The improvements on the TED (the bars you made a few weeks ago.)

To validate the optimization of ARB on complex character recognition and long-tailed character recognition, we divide them into different groups according to their complexity and frequency. 
%For each category, the differences in recognizing performances are first analyzed. 
For each category, to give a better understanding of the differences in recognizing performances, the character prediction samples by SAN and ABINet are studied in more detail.
%To give a better understanding of the performance differences, the misrecognized characters by at least one model (misrecognized by SAN, ABINet, or both) are studied in more detail.
%Specifically, the accuracy and the average TreeSim between the ground truth and character prediction samples are computed as a rough indicator of structural information awareness.
Specifically, we calculated the accuracy of character prediction results and the average TreeSim between the predicted results and the ground truth. TreeSim serves as a more fine-grained metric compared to accuracy, as it can indicate the awareness of structural information.

%To validate the optimization of ARB on complex character recognition and long-tailed character recognition,
%we select the misrecognized characters of SAN and ABINet, divide them into different groups according to their complexity and frequency, and investigate the difference on average TreeSim.

%To validate the optimization of ARB on complex character recognition and long-tailed character recognition, we select the both misrecognized characters of ARB-TreeSim and ABINet, divide them into different groups according to their complexity and frequency, and investigate the difference on average TreeSim.

\subsubsection{Experiments on different complexity characters}

\begin{figure}[h]	
\centering
    \subfigure[Web Dataset]
    {
        %\begin{minipage}[t]{0.43\linewidth}
        %    \centering          %子图居中
        %    \includegraphics[height=3.5cm,width=5cm]{figures/RSSL-web1-all.png}%{figures/web-radicalLen1-complex.png}   
            %\caption{Radical Tree}
        %\end{minipage}
        %\begin{minipage}[t]{0.43\linewidth}
        %    \centering      %子图居中
        %    \includegraphics[height=3.5cm,width=5cm]{figures/RSSL-web2-all.png}%{figures/web-radicalLen2-complex.png}
            %\caption{TreeSim Weight}
        %\end{minipage}

        \begin{minipage}[t]{1\linewidth}
            \centering          %子图居中
            \includegraphics[height=3.4cm,width=12.3cm]{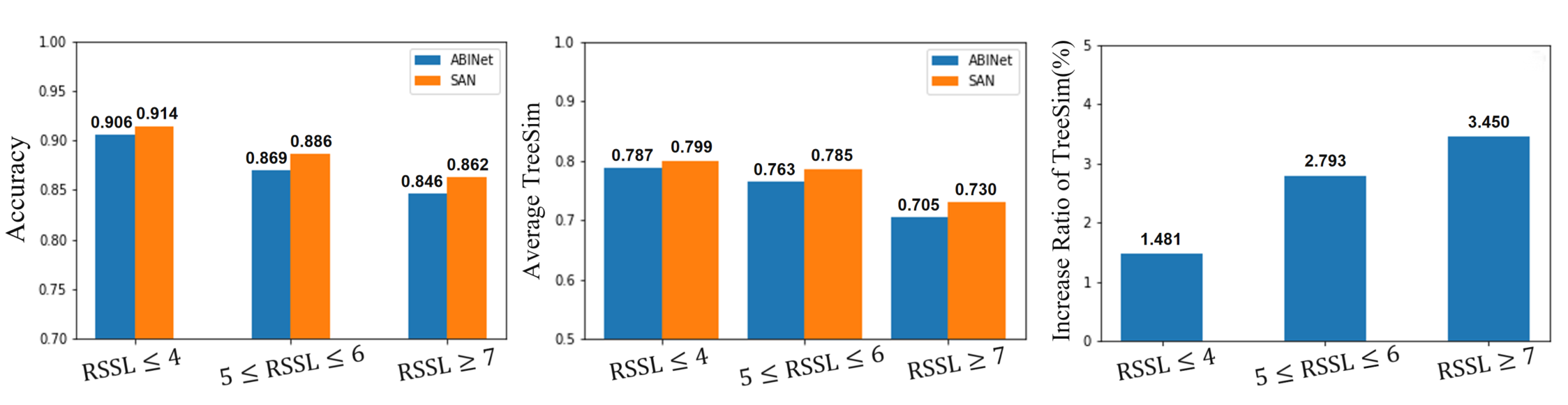}%{figures/web-radicalLen1-complex.png}   
            %\caption{Radical Tree}
        \end{minipage}
    }
    \subfigure[Scene Dataset]
    {
        %\begin{minipage}[t]{0.43\linewidth}
        %    \centering          %子图居中
        %    \includegraphics[height=3.5cm,width=5cm]{figures/RSSL-scene1-all.png}%{figures/scene-radicalLen1-complex.png}   
            %\caption{Radical Tree}
        %\end{minipage}
        %\begin{minipage}[t]{0.43\linewidth}
        %    \centering      %子图居中
        %    \includegraphics[height=3.5cm,width=5cm]{figures/RSSL-scene2-all.png}%{figures/scene-radicalLen2-complex.png}
            %\caption{TreeSim Weight}
        %\end{minipage}

        \begin{minipage}[t]{1\linewidth}
            \centering          %子图居中
            \includegraphics[height=3.4cm,width=12.3cm]{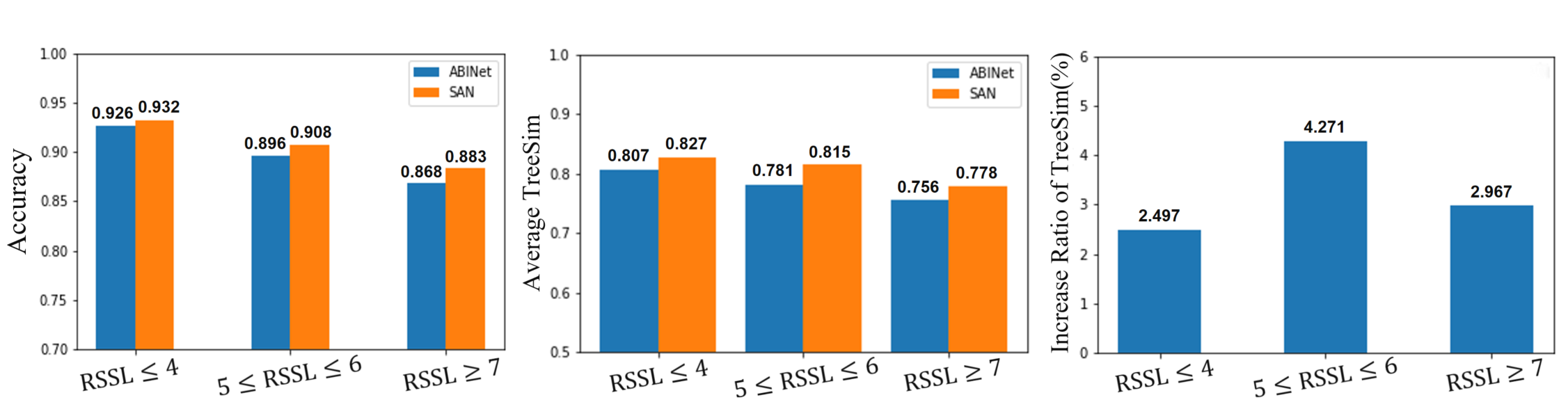}%{figures/web-radicalLen1-complex.png}   
            %\caption{Radical Tree}
        \end{minipage}
    }
\caption{The accuracy (left), the average TreeSim (middle) and the increase ratio of TreeSim (right) divided by RSSL on character prediction samples by SAN and ABINet. RSSL represents the length of the radical structure sequence of a character.}

%\vspace{-3mm}
\end{figure}

We observe that the complexity of the character structure is proportional to its radical structure sequence length~(RSSL), so we use RSSL to represent the complexity of each character. 
%Besides, we notice that the complexity of RSSL5(characters that RSSL is equal to 5) and RSSL6 is medium,
According to observations, we denote characters with complexity of RSSL5 (characters that RSSL is equal to 5) and RSSL6 as medium complexity characters,
characters with longer RSSL are denoted as complex and characters with shorter RSSL are considered as simple. 
Accodingly, we divide character classes into three parts: RSSL$\leq{}$4~(average 34$\%$ in web and 30$\%$ in scene), 5$\leq{}$RSSL$\leq{}$6~(average 38$\%$ in web and 37$\%$ in scene), RSSL$\ge$7~(average 28$\%$ in web and 33$\%$ in scene), and call them simple characters, sub-complex characters, and complex characters respectively.

%Besides, we notice that most RSSL is equal to 3, 5, and 7.
%We also notice that the complexity of RSSL5(characters that RSSL is equal to five) is medium, characters with longer RSSL are often complex and characters with shorter RSSL are simple. 
%At the same time, these three parts also have a similar quantity of characters.
%Hence we use RSSL5 as the dividing line and divide misrecognized characters into three parts: RSSL below five(average 35$\%$ in web and 31$\%$ in scene), RSSL equal to five(average 35$\%$ in web and 35$\%$ in scene) and RSSL above five(average 30$\%$ in web and 34$\%$ in scene), which we call them simple characters, sub-complex characters, and complex characters.

We calculate the accuracy and the average TreeSim between character prediction samples and ground truth values within each of these three parts~(Fig.5).
%From the result, we can see that for both web~(Fig.5 (a)) and scene dataset~(Fig.5 (b)), the average TreeSim in all three parts are all increasing.
%In the web dataset, the rising trend increases when RSSL increase, and the rise of complex characters~(RSSL$\ge{}$7) is more obvious.
%In scene dataset, the rise of sub-complex characters~(5$\leq{}$RSSL$\leq{}$6) is extremely significant and the increase of sub-complex and complex characters are both higher than simple characters~(RSSL$\leq{}$4).
%The experiments show that ARB can prominently improve the similarity between the recognition results and ground trues of complex characters, indicating that ARB has a more distinct perception of complex characters, 
%which is due to the introduction of hierarchical structure information improving the \textawareness{} of the vision model and making it easier for models to distinguish different components in complex characters.
%\textcolor{blue}{In terms of accuracy, we observe that for both the web dataset~(Fig.5 (a)) and the scene dataset~(Fig.5 (b)), the improvement in accuracy increases as the growth of RSSL progresses.}
%\textcolor{blue}{In terms of accuracy, we observe that for both the web dataset (Fig. 5(a)) and the scene dataset (Fig. 5(b)), the improvement in accuracy for complex~(RSSL$\ge{}$7) and sub-complex~(5$\leq{}$RSSL$\leq{}$6) characters is consistently higher than that for simple characters~(RSSL$\leq{}$4).}
In terms of accuracy, we note that in both the web dataset~(Fig. 5(a)) and the scene dataset~(Fig. 5(b)), there is a consistent trend of higher improvement in accuracy for sub-complex~(5$\leq{}$RSSL$\leq{}$6) and complex~(RSSL$\ge{}$7) characters compared to simple characters~(RSSL$\leq{}$4).
Regarding TreeSim, the average TreeSim in all three parts consistently increases. In the web dataset, the rising trend becomes more pronounced as RSSL increases, with the growth of complex characters being particularly noticeable. In the scene dataset, the increase in sub-complex characters is extremely significant, and the growth rates of both sub-complex and complex characters surpass that of simple characters.

The experiments show that ARB can prominently improve the similarity between the recognition results and ground trues of complex characters, indicating that ARB has a more distinct perception of complex characters, 
which is due to the introduction of hierarchical structure information improving the \textawareness{} of the vision model and making it easier for models to distinguish different components in complex characters.

\subsubsection{Experiments on long-tailed characters}

\begin{figure}[h]	
\centering
    \subfigure[Web Dataset]
    {
        %\begin{minipage}[t]{0.43\linewidth}
        %    \centering          %子图居中
        %    \includegraphics[height=3.5cm,width=5cm]{figures/OccN-web1-all.png}%{figures/web-frequency1-tail2.png}   
            %\caption{Radical Tree}
        %\end{minipage}
        %\begin{minipage}[t]{0.43\linewidth}
        %    \centering      %子图居中
        %    \includegraphics[height=3.5cm,width=4.7cm]{figures/OccN-web2-all.png}%{figures/web-frequency2-tail2.png}
            %\caption{TreeSim Weight}
        %\end{minipage}

        \begin{minipage}[t]{1\linewidth}
            \centering      %子图居中
            \includegraphics[height=3.4cm,width=12.3cm]{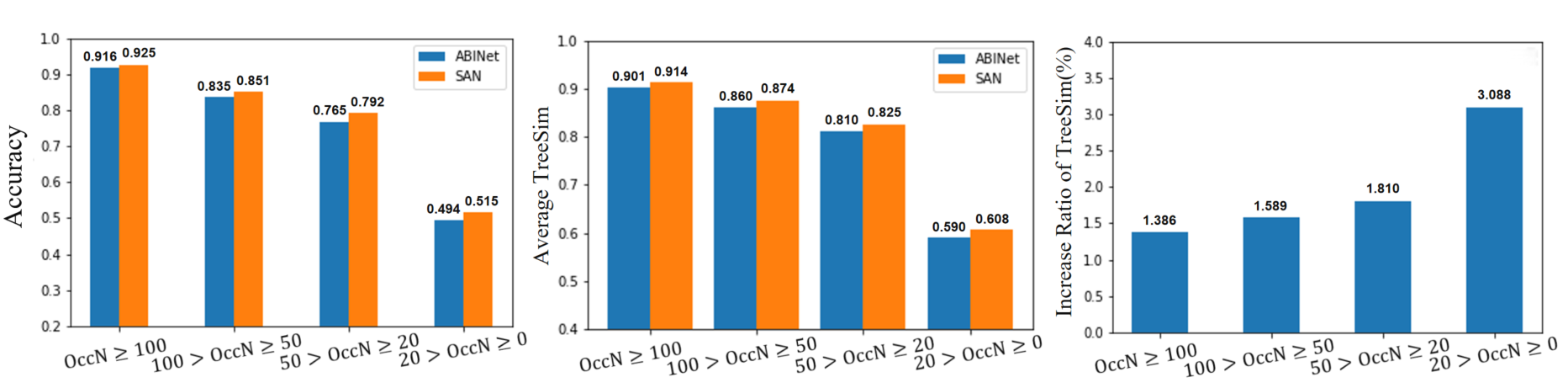}%{figures/web-frequency2-tail2.png}
            %\caption{TreeSim Weight}
        \end{minipage}
    }
    \subfigure[Scene Dataset]
    {
        %\begin{minipage}[t]{0.43\linewidth}
        %    \centering          %子图居中
        %    \includegraphics[height=3.5cm,width=5cm]{figures/OccN-scene1-all.png}%{figures/scene-frequency1-tail2.png}   
            %\caption{Radical Tree}
        %\end{minipage}
        %\begin{minipage}[t]{0.43\linewidth}
        %    \centering      %子图居中
        %    \includegraphics[height=3.5cm,width=5cm]{figures/OccN-scene2-all.png}%{figures/scene-frequency2-tail2.png}
            %\caption{TreeSim Weight}
        %\end{minipage}

        \begin{minipage}[t]{1\linewidth}
            \centering      %子图居中
            \includegraphics[height=3.4cm,width=12.3cm]{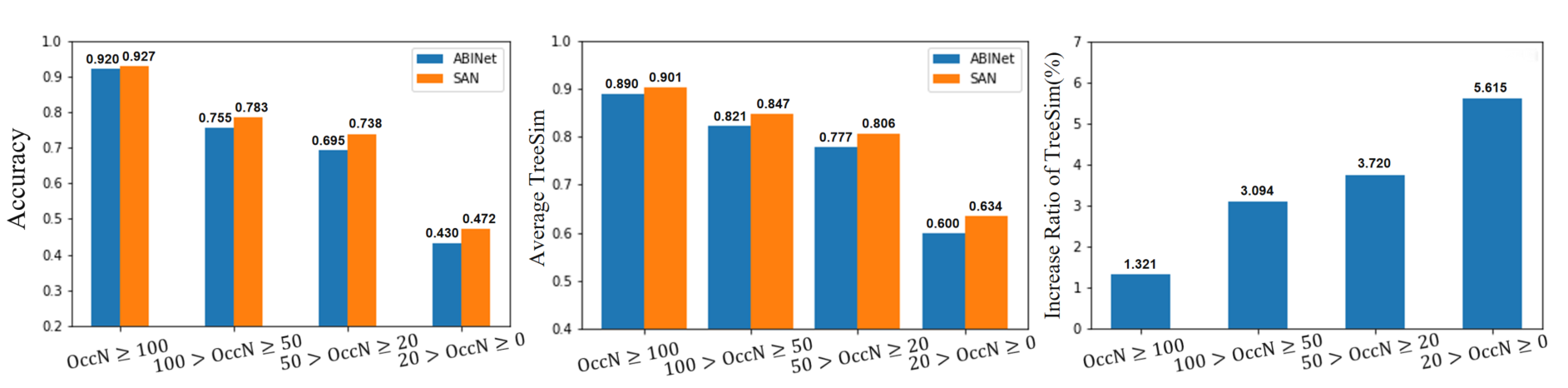}%{figures/web-frequency2-tail2.png}
            %\caption{TreeSim Weight}
        \end{minipage}
    }
%\caption{Average TreeSim divided by OccN on character prediction samples by SAN and ABINet. OccN is the occurrence number of a character in the training dataset. }
\caption{The accuracy (left), the average TreeSim (middle) and the increase ratio of TreeSim (right) divided by OccN on character prediction samples by SAN and ABINet. OccN represents the occurrence number of a character in the training dataset. }

\end{figure}

To validate the feasibility of our model on long-tailed characters, we sort character prediction samples in descending according to their occurrence number~(OccN) in the training dataset, 
and divide them into four parts: OccN$\ge$100~(average 22$\%$ in web and 39$\%$ in scene), 100$>$OccN$\ge$50~(average 12$\%$ in web and 10$\%$ in scene), 50$>$OccN$\ge$20~(average 19$\%$ in web and 15$\%$ in scene), 20$>$OccN$\ge$0~(average 47$\%$ in web and 36$\%$ in scene).

%To validate the optimation on long-tailed characters, we sort both misrecognized characters in descending according to their occurrence number in the training dataset, then divide them into four parts according to their frequency. The first part is the characters in the first 25$\%$ of the frequency, which is called head classes, and the last part is the characters in the last 25$\%$ of the frequency, which is called tail classes.

We calculate the accuracy and the average TreeSim between the character prediction samples and their corresponding ground truth within each of these parts~(Fig.6).
In terms of accuracy, we observe that for both the web dataset (Fig. 6(a)) and the scene dataset (Fig. 6(b)), the improvement in accuracy for infrequent characters~(OccN$<$50) is consistently higher than that for frequent characters~(OccN$\ge$50).
%The results show that in both web~(Fig.6 (a)) and scene dataset~(Fig.6 (b)), the rising trend is increasing when the occurrence number decreases, and the average TreeSim of tail classes(20$>$OccN$\ge$0) increases most obviously in both dataset.
%Regarding TreeSim, the rising trend is increasing when the occurrence number decreases, and the average TreeSim of tail classes~(20$>$OccN$\ge$0) increases most obviously in both dataset.
Regarding TreeSim, the rising trend becomes more pronounced as the occurrence number decreases. The average TreeSim of tail classes~(20$>$OccN$\ge$0) exhibits the most noticeable increase in both datasets.

These results demonstrate that the ARB can make the recognition results of long-tailed characters more similar to the ground truths, 
indicating that ARB can learn more features of long-tailed characters,
which is because of the common components combination traits shared between characters, ARB can take advantage of the traits learning on head classes to optimize the tail classes recognition.

%We calculate the average TreeSim between misrecognized characters and ground trues in these parts(Fig.7).
%The result shows that there are obviously decreasing in average TreeSim on tail classes, and the decline is significantly higher than other parts in both web and scene dataset.
%This result indicates that ARB has significant improvement in the recognition of long-tailed characters, which is because of the common components combination traits shared between characters, ARB can take advantage of the traits learning on head classes to optimize the tail classes recognition.

%\section{Limitation}

\section{Conclusion}

In this paper, we propose a Structure-Aware Network~(SAN) to optimize the recognition performance of complex characters and long-tailed characters, by using the proposed Auxiliary Radical Branch~(ARB) which utilizes the hierarchical components information of characters. 
Besides, we propose using Tree Similarity~(TreeSim) to measure the similarity of two characters and using TreeSim weight to enhance the structural information of label representation. 
Experiment results demonstrate the superiority of ARB on complex and long-tailed character recognition and validate that our method outperforms standard benchmarks and recent SOTA works.

\section{Acknowledgement}

The research is supported by National Key Research and Development Program of China (2020AAA0109700), National Science Fund for Distinguished Young Scholars (62125601), National Natural Science Foundation of China (62076024, 62006018), Interdisciplinary Research Project for Young Teachers of USTB (Fundamental Research Funds for the Central Universities)(FRF-IDRY-21-018).

\bibliographystyle{splncs04}\bibliography{cvpr}

\begin{thebibliography}{10}
\providecommand{\url}[1]{\texttt{#1}}
\providecommand{\urlprefix}{URL }
\providecommand{\doi}[1]{https://doi.org/#1}

\bibitem{www}
Baek, J., Kim, G., Lee, J., Park, S., Han, D., Yun, S., Oh, S.J., Lee, H.: What
  is wrong with scene text recognition model comparisons? dataset and model
  analysis. In: ICCV. pp. 4714--4722 (2019)

\bibitem{ZeroshotHC}
Cao, Z., Lu, J., Cui, S., Zhang, C.: Zero-shot handwritten chinese character
  recognition with hierarchical decomposition embedding. Pattern Recognit.
  \textbf{107},  107488 (2020)

\bibitem{DBLP:conf/icfhr/ChandaBHHSS18}
Chanda, S., Baas, J., Haitink, D., Hamel, S., Stutzmann, D., Schomaker, L.:
  Zero-shot learning based approach for medieval word recognition using
  deep-learned features. In: 16th International Conference on Frontiers in
  Handwriting Recognition, {ICFHR} 2018, Niagara Falls, NY, USA, August 5-8,
  2018. pp. 345--350. {IEEE} Computer Society (2018).
  \doi{10.1109/ICFHR-2018.2018.00067},
  \url{https://doi.org/10.1109/ICFHR-2018.2018.00067}

\bibitem{9412607}
Chanda, S., Haitink, D., Prasad, P.K., Baas, J., Pal, U., Schomaker, L.:
  Recognizing bengali word images - a zero-shot learning perspective. In: 2020
  25th International Conference on Pattern Recognition (ICPR). pp. 5603--5610
  (2021). \doi{10.1109/ICPR48806.2021.9412607}

\bibitem{Chen2021SceneTT}
Chen, J., Li, B., Xue, X.: Scene text telescope: Text-focused scene image
  super-resolution. 2021 IEEE/CVF Conference on Computer Vision and Pattern
  Recognition (CVPR) pp. 12021--12030 (2021)

\bibitem{taktak}
Chen, J., Li, B., Xue, X.: Zero-shot chinese character recognition with
  stroke-level decomposition. In: IJCAI. pp. 615--621 (2021)

\bibitem{fudanvi}
Chen, J., Yu, H., Ma, J., Guan, M., Xu, X., Wang, X., Qu, S., Li, B., Xue, X.:
  Benchmarking chinese text recognition: Datasets, baselines, and an empirical
  study. arXiv preprint arXiv:2112.15093  (2021)

\bibitem{MaximumER}
Cheng, C., Xu, W., Bai, X., Feng, B., Liu, W.: Maximum entropy regularization
  and chinese text recognition. ArXiv  \textbf{abs/2007.04651} (2020)

\bibitem{MultiColumnDN}
Ciresan, D.C., Meier, U.: Multi-column deep neural networks for offline
  handwritten chinese character classification. 2015 International Joint
  Conference on Neural Networks (IJCNN) pp.~1--6 (2013)

\bibitem{DeepNN}
Du, J., Wang, Z., Zhai, J.F., Hu, J.: Deep neural network based hidden markov
  model for offline handwritten chinese text recognition. 2016 23rd
  International Conference on Pattern Recognition (ICPR) pp. 3428--3433 (2016)

\bibitem{ABINet}
Fang, S., Xie, H., Wang, Y., Mao, Z., Zhang, Y.: Read like humans: Autonomous,
  bidirectional and iterative language modeling for scene text recognition. In:
  CVPR. pp. 7098--7107 (2021)

\bibitem{wubizhengma}
He, S., Schomaker, L.: Open set chinese character recognition using multi-typed
  attributes. arXiv preprint arXiv:1808.08993  (2018)

\bibitem{SAR}
Li, H., Wang, P., Shen, C., Zhang, G.: Show, attend and read: {A} simple and
  strong baseline for irregular text recognition. In: AAAI. pp. 8610--8617
  (2019)

\bibitem{moran}
Luo, C., Jin, L., Sun, Z.: {MORAN:} {A} multi-object rectified attention
  network for scene text recognition. Pattern Recognition  \textbf{90},
  109--118 (2019)

\bibitem{seed}
Qiao, Z., Zhou, Y., Yang, D., Zhou, Y., Wang, W.: {SEED:} semantics enhanced
  encoder-decoder framework for scene text recognition. In: CVPR. pp.
  13525--13534 (2020)

\bibitem{Rai2021PhoSCNetAA}
Rai, A., Krishnan, N.C., Chanda, S.: Pho(sc)net: An approach towards zero-shot
  word image recognition in historical documents. ArXiv
  \textbf{abs/2105.15093} (2021)

\bibitem{CRNN}
Shi, B., Bai, X., Yao, C.: An end-to-end trainable neural network for
  image-based sequence recognition and its application to scene text
  recognition. {IEEE} Trans. Pattern Anal. Mach. Intell.  \textbf{39}(11),
  2298--2304 (2017)

\bibitem{ASTER}
Shi, B., Yang, M., Wang, X., Lyu, P., Yao, C., Bai, X.: {ASTER:} an attentional
  scene text recognizer with flexible rectification. {IEEE} Trans. Pattern
  Anal. Mach. Intell.  \textbf{41}(9),  2035--2048 (2019)

\bibitem{HandwrittenCT}
Wang, Q.F., Yin, F., Liu, C.L.: Handwritten chinese text recognition by
  integrating multiple contexts. IEEE Transactions on Pattern Analysis and
  Machine Intelligence  \textbf{34},  1469--1481 (2012)

\bibitem{RadicalAN}
Wang, T., Xie, Z., Li, Z., Jin, L., Chen, X.: Radical aggregation network for
  few-shot offline handwritten chinese character recognition. Pattern Recognit.
  Lett.  \textbf{125},  821--827 (2019)

\bibitem{DenseRANFO}
Wang, W., shu Zhang, J., Du, J., Wang, Z., Zhu, Y.: Denseran for offline
  handwritten chinese character recognition. 2018 16th International Conference
  on Frontiers in Handwriting Recognition (ICFHR) pp. 104--109 (2018)

\bibitem{JointAA}
Wang, Z., Du, J.: Joint architecture and knowledge distillation in cnn for
  chinese text recognition. Pattern Recognit.  \textbf{111},  107722 (2019)

\bibitem{WriterAwareCF}
Wang, Z., Du, J., Wang, J.: Writer-aware cnn for parsimonious hmm-based offline
  handwritten chinese text recognition. ArXiv  \textbf{abs/1812.09809} (2018)

\bibitem{JointSA}
Wu, C.J., Wang, Z., Du, J., shu Zhang, J., Wang, J.: Joint spatial and radical
  analysis network for distorted chinese character recognition. 2019
  International Conference on Document Analysis and Recognition Workshops
  (ICDARW)  \textbf{5},  122--127 (2019)

\bibitem{HandwrittenCR}
Wu, C., liang Fan, W., He, Y., Sun, J., Naoi, S.: Handwritten character
  recognition by alternately trained relaxation convolutional neural network.
  2014 14th International Conference on Frontiers in Handwriting Recognition
  pp. 291--296 (2014)

\bibitem{FromTT}
Wu, Y., Hu, X.: From textline to paragraph: A promising practice for chinese
  text recognition. In: Proceedings of the Future Technologies Conference. pp.
  618--633. Springer (2020)

\bibitem{BuildingFA}
Xiao, X., Jin, L., Yang, Y., Yang, W., Sun, J., Chang, T.: Building fast and
  compact convolutional neural networks for offline handwritten chinese
  character recognition. Pattern Recognit.  \textbf{72},  72--81 (2017)

\bibitem{TemplateInstanceLF}
Xiao, Y., Meng, D., Lu, C., Tang, C.K.: Template-instance loss for offline
  handwritten chinese character recognition. 2019 International Conference on
  Document Analysis and Recognition (ICDAR) pp. 315--322 (2019)

\bibitem{ICDAR2C}
Yin, F., Wang, Q.F., Zhang, X.Y., Liu, C.L.: Icdar 2013 chinese handwriting
  recognition competition. 2013 12th International Conference on Document
  Analysis and Recognition pp. 1464--1470 (2013)

\bibitem{SRN}
Yu, D., Li, X., Zhang, C., Liu, T., Han, J., Liu, J., Ding, E.: Towards
  accurate scene text recognition with semantic reasoning networks. In: CVPR.
  pp. 12110--12119 (2020)

\bibitem{RANRA}
shu Zhang, J., Zhu, Y., Du, J., Dai, L.: Ran: Radical analysis networks for
  zero-shot learning of chinese characters. ArXiv  \textbf{abs/1711.01889}
  (2017)

\end{thebibliography}

\end{document}